\documentclass[letterpaper,twocolumn,10pt]{article}
\usepackage{usenix-2020-09}

\usepackage{tikz}
\usepackage{amsmath}
\usepackage{amssymb}
\usepackage{subcaption}
\usepackage{algorithm}
\usepackage[noend]{algpseudocode}
\usepackage{url}
\usepackage{multirow}
\usepackage{pifont}
\newcommand{\sys}{\textsc{Comb}}

\begin{document}

\date{}

\title{\Large \bf You Need an Encoder for Native Position-Independent Caching}

\author{
Shiju Zhao$^*$, Junhao Hu$^\dag$, Jiaqi Zheng$^*$, Guihai Chen$^*$\\
$^*$State Key Laboratory for Novel Software Technology, Nanjing University, China\\
$^\dag$School of Computer Science, Peking University, China
}

\maketitle

\begin{abstract}
The Key-Value (KV) cache of Large Language Models (LLMs) is prefix-based, making it highly inefficient for processing contexts retrieved in arbitrary order. Position-Independent Caching (PIC) has been proposed to enable KV reuse without positional constraints; however, existing approaches often incur substantial accuracy degradation, limiting their practical adoption. To address this issue, we propose native PIC by reintroducing the encoder to prevalent decoder-only LLMs and explicitly training it to support PIC. We further develop \sys, a PIC-aware caching system that integrates seamlessly with existing inference frameworks. Experimental results show that \sys\ reduces Time-to-First-Token (TTFT) by 51-94\% and increases throughput by 3$\times$ with comparable accuracy. Furthermore, the quality improvement when using DeepSeek-V2-Lite-Chat demonstrates the applicability of \sys\ to other types of decoder-only LLMs. Our code is available at https://github.com/shijuzhao/Comb.
\end{abstract}

\section{Introduction}\label{sec-intro}

Large Language Models (LLMs) exhibit strong capabilities across a wide range of complex tasks, such as automatic software engineering~\cite{wang2023how, hu2023pcrml, liang2025diff}, document-question answering~\cite{bai2024longbench}, and reasoning-intensive problems~\cite{yao2023react}. LLMs also offer a user-friendly interface through natural-language-based prompts---sequences of tokens. As LLMs' capabilities and ease of use expand, their usage patterns have shifted from simple conversational tasks to more complex scenarios, such as multi-turn planning, reasoning, tool use, and few-shot learning. This shift has resulted in long prompts with repeated tokens across requests, such as system messages, few-shot learning examples, and documents, whose content is less changing (i.e., static) than user-specific instructions (being dynamic).

\begin{figure}[t]
    \centering
    \includegraphics[width=\linewidth]{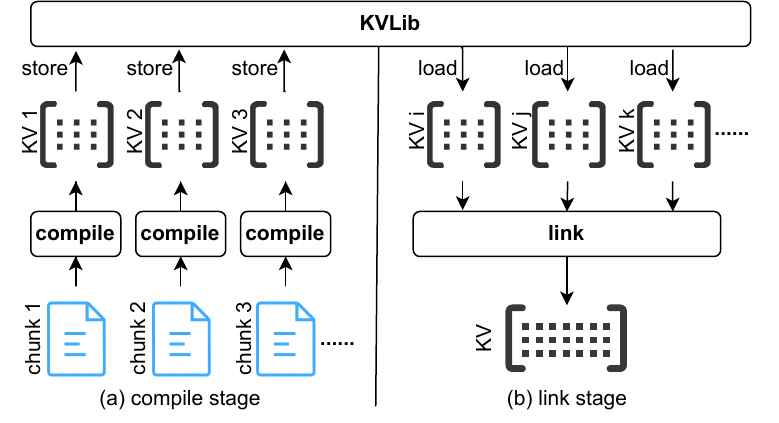}
    \caption{Position-Independent Caching (PIC) workflow. $KV_i$, $KV_j$, $KV_k$ indicate KV vectors in arbitrary order/position.}
    \label{fig-pic}
\end{figure}

Prefix-Based Context Caching (Prefix caching) is an efficient approach that reuses Key-Value (KV) vectors, the intermediate representations of repeated tokens in previous requests, to reduce computation. Prefix caching matches the current request against previous ones to reuse the KV vectors of the longest common prefix. Prefix caching remains the dominant approach in existing systems~\cite{deepseek_api, gemini-context-caching, zheng2024sglang, kwon2023vllm}, but it requires exact prefix matches across requests, limiting reuse cases in settings such as few-shot learning and Retrieval-Augmented Generation (RAG), where static chunks (e.g., documents) remain unchanged across requests but are preceded by varying prefixes. 

To address the limitations of prefix caching, Position-Independent Caching (PIC)~\cite{yao2025cacheblend, hu2025epic} enables modular reuse of the KV vectors of static tokens, regardless of their prefixes. Specifically, the PIC technique is a two-stage framework analogous to compilation and linking (Figure~\ref{fig-pic}). First, the \textbf{compile} stage involves submitting individual static chunks to the LLM to generate and store their respective KV vectors from position zero. Second, the \textbf{link} stage loads and concatenates needed KV vectors (in any order/position), with optional computations to recover accuracy. PIC significantly increases reuse opportunities, but it deviates from standard attention mechanisms, resulting in potential accuracy degradation; thus, ensuring accurate recovery becomes its main challenge.

Subsequent studies largely conform to the PIC framework and can be categorized into two paradigms: \textbf{post-training PIC} and \textbf{training-aware PIC}. \textbf{Post-training PIC}, typically recompute a subset of tokens in the link stage to recover accuracy~\cite{hu2025epic, yao2025cacheblend, agarwal2025cachecraft, zhang2025attention, zhao2025mpic}. These approaches are non-intrusive (they do not train the model or modify model architectures) and can be deployed as plug-ins, but they generally suffer from reduced accuracy. \textbf{Training-aware} PIC explicitly trains models to be aware of the PIC usage during the compile stage, thereby achieving near-zero overhead in the link stage~\cite{yang2025kvlink, ma2025blockattention, lu2025turborag}. These approaches achieve high accuracy, but they require fine-tuning and thus permanently alter model behavior, which may lead to catastrophic forgetting and performance degradation on other tasks (Figure~\ref{fig:block}).

To address these limitations and combine the advantages of both paradigms, we propose \sys, which reintroduces an encoder plug-in into mainstream decoder-only LLMs and trains only the encoder to be PIC-aware. This design offers three key benefits. First, compared to post-training approaches, \sys\ achieves the highest accuracy by integrating a PIC-specific component---the encoder---and training it explicitly for PIC generation, thereby making PIC a native capability of the model, analogous to native sparse attention in DeepSeek~\cite{nsa2025yuan}. Second, compared to training-aware approaches, \sys\ provides maximal flexibility: the encoder functions as a plug-in whose cross-attention can be entirely removed without affecting the standard decoding workflow. Third, compared to all approaches, despite introducing additional parameters (from the encoder), \sys\ achieves the lowest Time-To-First-Token (TTFT) due to the efficiency of \sys's encoder-decoder architecture (Section~\ref{sec-model}).

We implement \sys\ with two components: a model component and a system component. For the model component, we reintroduce an encoder into standard decoder-only architectures, freeze the decoder parameters, and explicitly train the encoder to accommodate PIC. Input chunks are independently processed by the encoder to generate their KV vectors, and queries over these chunks are supervised using a maximum-likelihood objective against ground-truth outputs. Architecturally, the encoder layers are interleaved with the decoder in a comb-like manner and perform cross-attention only, motivating the name of \sys. For the system component, we implement a PIC management system on top of the existing inference framework, including HuggingFace transformers and vLLM~\cite{kwon2023vllm}, comprising approximately 5K lines of Python code. The code and models that we trained are publicly available online and have been anonymized.

Experimental results on LongBench~\cite{bai2024longbench} show that \sys\ achieves the highest accuracy and the lowest TTFT among all evaluated approaches. When cache hits occur, \sys\ reduces TTFT by up to 94\% and improves throughput by up to $3\times$, while matching or exceeding the accuracy of prefix-based attention. Importantly, these gains are achieved without permanently modifying the underlying decoder-only model: \sys\ remains a plug-in mechanism that can be enabled or disabled on demand, reverting to standard prefix caching when PIC is not used. Furthermore, \sys\ saves memory usage for KV vectors by 75\%.
\section{Background}\label{sec-background}

This section provides a primer on transformers, context caching, and its variant, Position-Independent Caching (PIC), along with a review of existing PIC approaches.

\subsection{Autoregressive Generation and KV Cache}
The generation process of Large Language Models (LLMs) consists of two distinct stages: the prefill stage and the decode stage. In the \textbf{prefill} stage, the model processes a sequence of prompt tokens all at once. It computes the KV vectors for all prompt tokens, stores these vectors in the KV cache, and generates the first output token to initiate the decode stage. The time required to generate the first token is referred to as the Time-To-First-Token (TTFT). In the \textbf{decode} stage, the model iteratively processes each newly generated token. It computes the KV vectors for the new token, appends these vectors to the KV cache, and generates the next token. This process repeats until a specified stopping criterion is met.
 
\subsection{Position-Independent Caching (PIC)} 

Context caching techniques can be broadly categorized into prefix-based caching and PIC. In this work, we focus on PIC, which enables the reuse of cached KVs of static prompt segments independent of their positions or surrounding context. Existing PIC approaches generally follow a two-stage framework analogous to compilation and linking (Figure~\ref{fig-pic}), as introduced in Section~\ref{sec-intro}.

Prior PIC approaches can be further classified into \emph{post-training PIC} and \emph{training-aware PIC}. \textbf{Post-training PIC} approaches, including EPIC~\cite{hu2025epic} and CacheBlend~\cite{yao2025cacheblend}, treat the underlying decoder-only model as fixed and recover accuracy through selective recomputation during the link stage. For example, EPIC recomputes initial tokens of each chunk~\cite{hu2025epic}, while CacheBlend recomputes a subset of tokens with the largest discrepancies~\cite{yao2025cacheblend}. \textbf{Training-aware PIC} approaches explicitly incorporate PIC into model training~\cite{yang2025kvlink, ma2025blockattention, lu2025turborag}, enabling near-zero-overhead reuse while maintaining high accuracy. For example, BlockAttention~\cite{ma2025blockattention} enforces PIC by training the model to restrict attention within document blocks, preventing cross-document interactions. KVLink~\cite{yang2025kvlink} introduces and trains special linking tokens that reestablish effective self-attention across independently cached documents.
\section{System Overview}
\begin{figure}
    \centering
    \includegraphics[width=\linewidth]{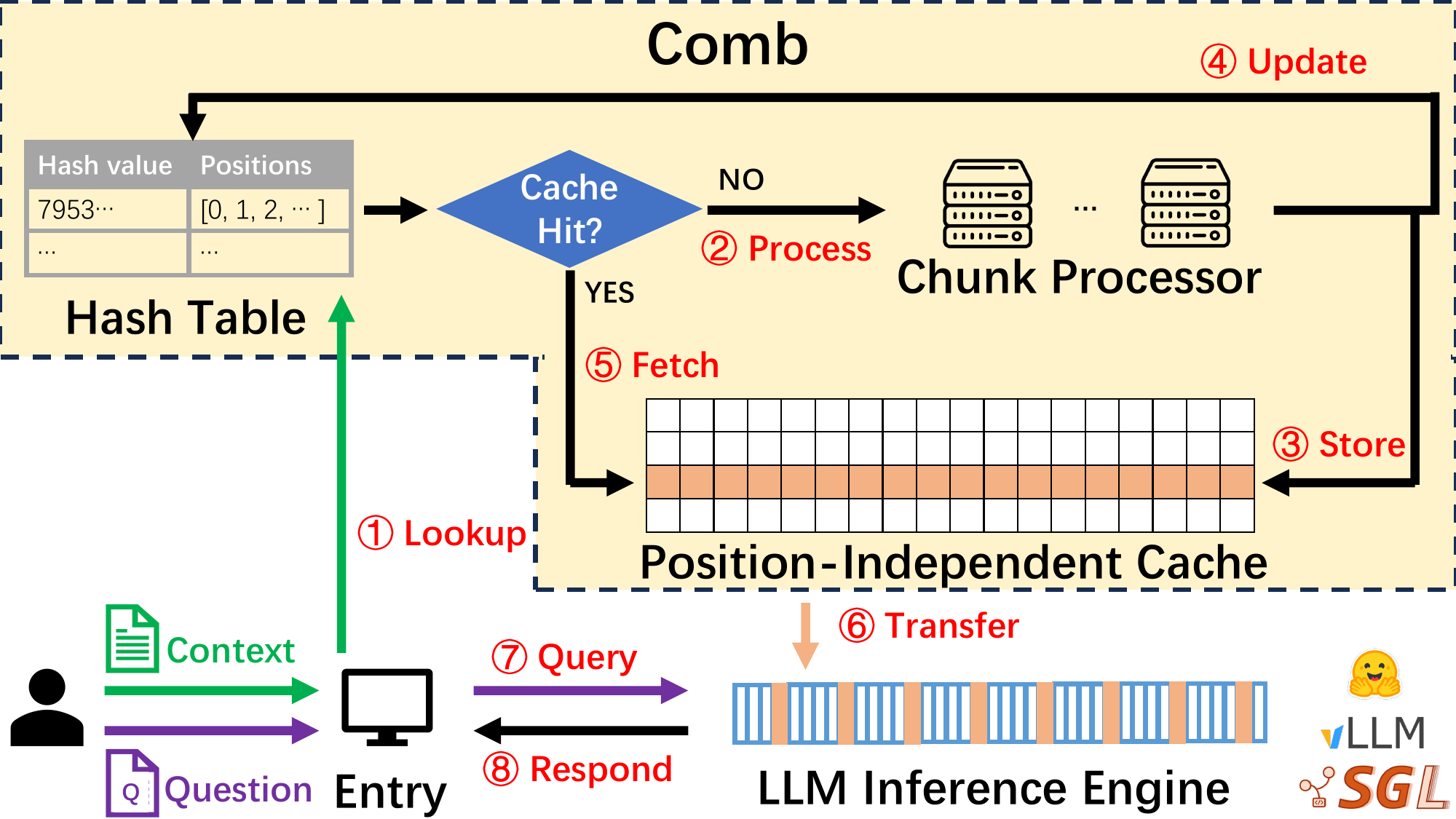}
    \caption{\sys\ Overview.}
    \label{fig-system}
\end{figure}

We start by elucidating the workflow of \sys\ serving system as shown in \figurename~\ref{fig-system}, and explain how we render the PIC intrinsic to the model architecture in the next section.

The input of \sys\ consists of a question and possibly none or multiple contexts, which is expected to be reused in an arbitrary position. We let users distinguish between context and query because users know their own needs best. \textcircled{1} For multiple contexts, \sys\ looks up whether their Position-Independent Caches (PICaches\footnote{In this paper, we use the term \emph{PIC} to refer to the position-independent caching technology, and the term \emph{PICache} to refer to the actual position-independent KV cache.}) exist through a hash table. \textcircled{2} For those contexts without caches, a chunk processor will generate their KV caches, \textcircled{3} store the KV caches, and \textcircled{4} update the hash table. Then the PICaches of these contexts are fetched \textcircled{5} and transferred to an LLM inference engine (e.g., HuggingFace transformers, vLLM, or SGLang) \textcircled{6}. \textcircled{7} The question is passed directly to the LLM inference engine, and is responded to with the PICaches \textcircled{8}. The blue and orange rectangles in the lower right corner of the figure represent the self-attention layers and cross-attention layers of LLM, which will be detailed in the next section. Note that \sys\ can be seamlessly integrated into the existing disaggregated prefill-decode serving paradigm. The chunk processor can be considered the prefill node, while the LLM inference engine can be considered the decode node.

\section{Model Design}\label{sec-model}

\begin{figure}
    \centering
    \includegraphics[width=\linewidth]{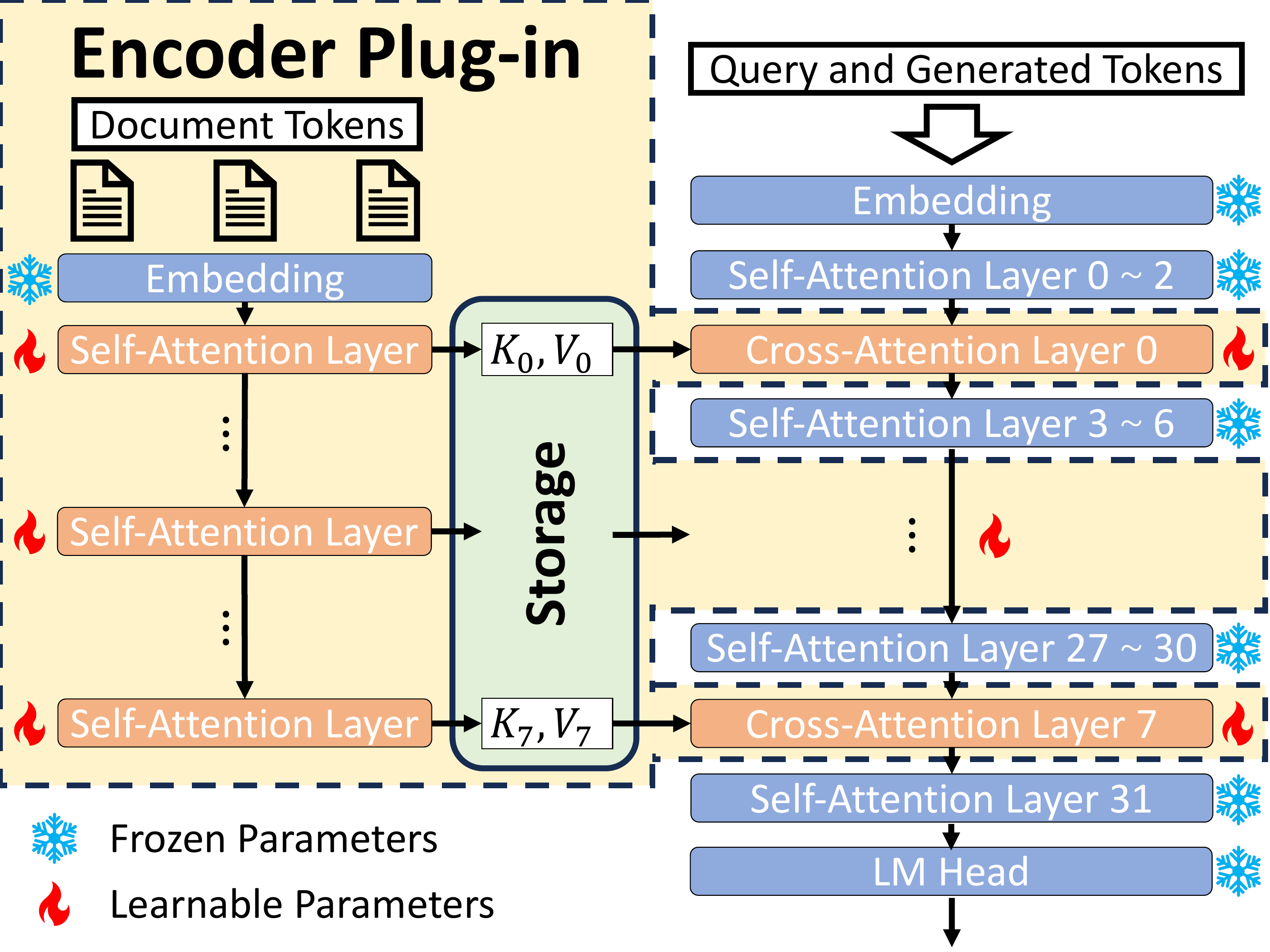}
    \caption{Model architecture.}
    \label{fig-model}
\end{figure}

We now describe the architecture of \sys\ and analyze its computational and memory complexity. As illustrated in \figurename~\ref{fig-model}, \sys\ resembles a classical encoder–decoder model but differs in key aspects. The decoder's goal is text generation. During decoding, newly generated tokens are appended to the query sequence and treated identically to the initial query tokens. The encoder is trained to support PIC. Since all decoder parameters are frozen during training, disabling the encoder and the cross-attention layers in \figurename~\ref{fig-model} recovers the original decoder-only LLM.

Throughout, we denote the hidden dimension by $d$, the document length by $l_{\text{doc}}$, and the query length by $l_{\text{query}}$. We present all attention computations in the single-head case; the extension to multi-head attention is standard and does not affect asymptotic complexity.

\subsection{Self-Attention Layer}

The decoder (blue self-attention layers in Figure~\ref{fig-model}) is taken directly from an existing decoder-only LLM that handles only queries and generated tokens. The encoder (orange self-attention layers in Figure~\ref{fig-model}) shares the same per-layer structure as the decoder (self-attention plus feed-forward), but has \emph{fewer} layers than the decoder (e.g., 8 encoder layers interleaved with 32 decoder layers in Figure~\ref{fig-model}). Its embedding layer inherits from the decoder and remains frozen. The encoder processes document tokens and produces PICaches. 

Both the encoder and decoder comprise multiple self-attention layers. Let the input of a self-attention layer be $X \in \mathbb{R}^{l_{\text{query}} \times d}$. In each layer, we first compute
\begin{equation}
    Q = X W_{Q}, K = X W_{K}, V = X W_{V},
\end{equation}
where $W_{Q}, W_{K}, W_{V} \in \mathbb{R}^{d \times d}$ are the query, key, and value projection matrices. Masked self-attention is given by
\begin{align}
    A &= \mathrm{softmax}\left(\frac{Q K^\top}{\sqrt{d}} + M\right), & A &\in \mathbb{R}^{l_{\text{query}} \times l_{\text{query}}}, \\
    H &= A V, & H &\in \mathbb{R}^{l_{\text{query}} \times d},
\end{align}
where $M \in \mathbb{R}^{l_{\text{query}} \times l_{\text{query}}}$ is the causal mask and $H$ is the obtained hidden states to be processed by the next module. The remaining feed-forward, residual, and normalization components also follow the standard decoder-only architecture. 

\subsection{Cross-Attention Layer}

Encoder--decoder interactions occur in the orange cross-attention layers in Figure~\ref{fig-model}. In these layers, decoder tokens (query tokens or newly generated tokens) attend to the encoder-produced document KVs. Let the output of an encoder's self-attention layer be $H^E\in\mathbb{R}^{l_\text{doc}\times d}$, and the output of the previous decoder's self-attention layer be $X^D\in\mathbb{R}^{l_\text{query}\times d}$. The query, key, and value are given by
\begin{align}
    &Q^\text{cross}=X^DW^\text{cross}_Q,& Q^\text{cross}\in\mathbb{R}^{l_\text{query}\times d},\\
    &K^\text{cross}=H^EW_K^\text{cross},& K^\text{cross}\in\mathbb{R}^{l_\text{doc}\times d},\\
    &V^\text{cross}=H^EW_V^\text{cross},& V^\text{cross}\in\mathbb{R}^{l_\text{doc}\times d},
\end{align}
where $W_{Q}^\text{cross}, W_{K}^\text{cross}, W_{V}^\text{cross} \in \mathbb{R}^{d \times d}$ are the query, key, value projection of the orange cross-attention in \figurename~\ref{fig-model}. Cross-attention then computes
\begin{align}
    A^{\text{cross}} &= \mathrm{softmax}\left(\frac{Q^{\text{cross}} (K^{\text{cross}})^\top}{\sqrt{d}}\right), & A^{\text{cross}} &\in \mathbb{R}^{l_{\text{query}} \times l_{\text{doc}}}, \\
    H^{\text{cross}} &= A^{\text{cross}} V^{\text{cross}}, & H^{\text{cross}} &\in \mathbb{R}^{l_{\text{query}} \times d}.
\end{align}

The cross-attention output $H^{\text{cross}}$ is passed through the standard decoder pathway via residual connections inside the decoder blocks.

There are two options to implement PIC: (1) store $H^E$, or (2) store $K^\text{cross}$ and $V^\text{cross}$. The first option saves storage space, but introduces computational overhead after the PICache is loaded. As a result, we store $(K^\text{cross},V^\text{cross})$ for efficiency in the current design. The KV vectors form the document-side representations which are stored and reused across requests by \sys. The PICaches of multiple documents can be generated independently from position zero during the compile stage and later concatenated along the $l_{\text{doc}}$ during inference (link stage). 

\subsection{Complexity Analysis}

We now compare the computational and memory complexity of \sys\ with a standard decoder-only architecture. Let $N = l_{\text{doc}} + l_{\text{query}}$ denote the total sequence length.

\paragraph{Time Complexity.} A standard decoder-only architecture with $L_{\mathrm{D}}$ layers (e.g., $L_{\mathrm{D}} = 32$) applies self-attention to all $N$ tokens in each layer, yielding per-layer complexity $O(N^{2})$ and total
\begin{equation}
    T_{\mathrm{dec\text{-}only}} = O\big(L_{\mathrm{D}} N^{2}\big) = O\big(L_{\mathrm{D}} (l_{\text{query}} + l_{\text{doc}})^{2}\big).
\end{equation}

In \sys, we use $L_{\mathrm{D}}$ decoder layers and $L_{\mathrm{E}}$ encoder layers (e.g., $L_{\mathrm{D}} = 32$, $L_{\mathrm{E}} = 8$ in Figure~\ref{fig-model}). The encoder attends only over document tokens, with per-layer cost $O(l_{\text{doc}}^{2})$ and total
\begin{equation}
    T_{\mathrm{E}} = O\big(L_{\mathrm{E}} \, l_{\text{doc}}^{2}\big).
\end{equation}
Decoder self-attention operates only on query tokens, with per-layer cost $O(l_{\text{query}}^{2})$ and total
\begin{equation}
    T_{\mathrm{D,self}} = O\big(L_{\mathrm{D}} \, l_{\text{query}}^{2}\big).
\end{equation}
Each cross-attention layer (orange blocks in Figure~\ref{fig-model}) mixes $l_{\text{query}}$ decoder queries with $l_{\text{doc}}$ document KVs, for per-layer cost $O(l_{\text{query}} \, l_{\text{doc}})$ and total
\begin{equation}
    T_{\mathrm{cross}} = O\big(L_{\mathrm{cross}} \, l_{\text{query}} \, l_{\text{doc}}\big),
\end{equation}
where $L_{\mathrm{cross}} \leq L_{\mathrm{D}}$ is the number of cross-attention layers (e.g., $L_{\mathrm{cross}} = 8$ interleaved layers in Figure~\ref{fig-model}). The overall attention cost of \sys\ is
\begin{equation}
    T_{\mathrm{comb}} = O\big(L_{\mathrm{E}} \, l_{\text{doc}}^{2} + L_{\mathrm{D}} \, l_{\text{query}}^{2} + L_{\mathrm{cross}} \, l_{\text{query}} \, l_{\text{doc}}\big).
\end{equation}

Since $L_E$ and $L_{\mathrm{cross}}$ are less than $L_D$, $T_{\mathrm{comb}} < T_{\mathrm{dec\text{-}only}}$.

Two things are worth noting. First, in practice, each document is independently prefilled by the encoder. When prefilling $m$ documents with lengths $l_{\text{doc},1}, \ldots, l_{\text{doc},m}$ such that $l_{\text{doc}} = \sum_{i=1}^{m} l_{\text{doc},i}$, the encoder prefill cost scales as $\sum_{i=1}^{m} l_{\text{doc},i}^{2}$ rather than $l_{\text{doc}}^{2}$, and we have $\sum_{i=1}^{m} l_{\text{doc},i}^{2} \le l_{\text{doc}}^{2}$. Second, modern inference engines already employ chunked prefilling for long prompts by processing them in smaller segments iteratively. As a result, the initial PIC compilation (``cold start'') of a new document typically aligns with this existing procedure and, in cache-miss scenarios, does not introduce noticeable overhead beyond that of the baseline system.

\paragraph{Memory Complexity.} We now compare the memory required for KV vectors. In the standard decoder-only architecture, each of the $L_{\mathrm{D}}$ layers stores KVs for all $N$ tokens, for total KV memory
\begin{equation}
    M_{\mathrm{dec\text{-}only}} = O\big(L_{\mathrm{D}} N d\big) = O\big(L_{\mathrm{D}} (l_{\text{query}} + l_{\text{doc}}) d\big).
\end{equation}

In \sys, encoder KV vectors are stored for document tokens across $L_{\mathrm{E}}$ layers, with memory
\begin{equation}
    M_{\mathrm{E}} = O\big(L_{\mathrm{E}} \, l_{\text{doc}} d\big).
\end{equation}
Decoder KV vectors are stored only for query tokens,
\begin{equation}
    M_{\mathrm{D}} = O\big(L_{\mathrm{D}} \, l_{\text{query}} d\big).
\end{equation}
Thus, the total KV memory of \sys\ is
\begin{equation}
    M_{\mathrm{comb}} = O\big(L_{\mathrm{E}} \, l_{\text{doc}} d + L_{\mathrm{D}} \, l_{\text{query}} d\big).
\end{equation}

Since $L_E<L_D$, $M_{\mathrm{comb}}< M_{\mathrm{dec\text{-}only}}.$

Finally, $L_{\mathrm{E}}$ can be chosen in the range $[1, L_{\mathrm{D}}]$. Larger $L_{\mathrm{E}}$ increases expressivity and potentially accuracy, while even $L_{\mathrm{E}} = L_{\mathrm{D}}$ does not exceed the time or memory complexity of a decoder-only model. In our experiments, we set $L_{\mathrm{E}} = 8$ to balance accuracy and computational resources (Section~\ref{sec-eval}). Moreover, under our limited GPU resources, this configuration avoids out-of-memory errors and accelerates training by enabling larger batch sizes due to reduced parameter counts, optimizer states, and activation memory.
\section{Implementation}\label{sec-impl}

This section describes how we instantiate, train, and deploy \sys. We first detail the model information, then summarize the training data and recipe, and finally discuss how \sys\ is integrated into a production serving system.

\subsection{Models}

We implement \sys\ by augmenting a frozen decoder-only LLM with a trainable encoder described in Section~\ref{sec-model}. First, the decoder is directly adopted from existing open-source models. Specifically, we use two open-source LLMs: Llama-3.1-8B-Instruct~\cite{llama3.1} and DeepSeek-V2-Lite-Chat\footnote{https://huggingface.co/deepseek-ai/DeepSeek-V2-Lite-Chat} as the decoder backbones. Llama represents the most widely used architecture in the current LLM ecosystem, whereas DeepSeek departs from the standard KV-cache design by employing Multi-head Latent Attention (MLA) for improved memory efficiency. Together, these models cover diverse architectural choices and training paradigms. Second, we integrate an encoder into the decoder via cross-attention layers, where only the encoder parameters and the query-side projections in the cross-attention layers are trainable. The resulting models are referred to as \textsc{CombLlama} and \textsc{CombDeepSeek}, respectively.

Table~\ref{tab-impl-config} summarizes the parameter counts and training overhead. We train all models using the AdamW optimizer with linear warmup followed by cosine decay. Unless otherwise stated, gradient clipping and weight decay follow standard practices for large-scale language model fine-tuning. Additional implementation details are provided in the released code.

\begin{table}[t]
    \centering
    \caption{Model size and training overhead for \sys.}
    \label{tab-impl-config}
    \begin{small}
      \begin{sc}
        \begin{tabular}{l|cccr}
          \hline
          Model Name & CombLlama & CombDeepseek \\
          \hline
          \# of Frozen & \multirow{2}{*}{8.0 B} & \multirow{2}{*}{15.7 B}\\
          Parameters &   &   \\
          \hline
          \# of Learnable & \multirow{2}{*}{3.5 B} & \multirow{2}{*}{3.5 B} \\
          Parameters & &   \\
          \hline
          GPU Hours & \multirow{2}{*}{2966} & \multirow{2}{*}{5402} \\
          (A100 * hour) &  &  \\
          \hline
        \end{tabular}
      \end{sc}
    \end{small}
\end{table}

\subsection{Training Data and Recipe}

\sys\ is trained to emulate the behavior of the underlying decoder-only model under Position-Independent Caching. We select SQuAD~\cite{rajpurkar2018know}, Natural-Instructions~\cite{mishra2022naturalinstructions}, XSum~\cite{narayan2018dont}, and Super-Natural-Instructions~\cite{Wang2022supernaturalinstructions} as training datasets. Each training example is constructed as a tuple $(D, Q, Y)$, where $D$ is a (possibly multi-document) static context, $Q$ is a user query or instruction, and $Y$ is the ground-truth response. During training, each document in $D$ is first independently encoded by the encoder to obtain document-side KVs $(K^{E}, V^{E})$, and the frozen decoder then generates outputs conditioned on $(K^{E}, V^{E})$ and query tokens. We optimize a token-level cross-entropy loss between the decoder outputs and the target sequence $Y$, using teacher forcing. 

The ground-truth sequence $Y$ is generated by Llama-3.1-8B-Instruct given $(D, Q)$ as input, resulting in on-policy training for the Llama-based model. For DeepSeek, whose native generations are of lower quality, we also adopt Llama-generated outputs as supervision, which substantially improves its accuracy (Figure~\ref{fig:deepseek-f1}). All models are trained on four NVIDIA A100 80GB GPUs with tensor parallelism set to 4.

\subsection{System Details}

We build \sys\ system on top of existing LLM inference frameworks (e.g., huggingface transformers and vLLM~\cite{kwon2023vllm}), with 5K lines of code in Python.

\sys\ consists of four key components: \textbf{PIC manager}, \textbf{PIC allocator}, \textbf{Chunk processor}, and \textbf{Inference engine}. When the system starts, the PIC manager reserves a portion (specified by the operator) of GPU memory, initializes the PIC allocator, and launches the Chunk processor and the Inference engine. After initialization, the PIC manager will serve users' requests, which consist of a query and several documents (possibly none or multiple). First, the documents will be looked up to check if their PICaches exist. If not, the PIC allocator will allocate free memory for their PICaches, and the Chunk processor computes the encoder part for them. Then the PICaches are stored asynchronously. After collecting the PICaches, the PIC manager will send their handle to the Inference engine through the CUDA IPC (Inter-Process Communication) API. This enables tensors to be passed in place between processes. Finally, the Inference engine generates a response using the query and the related PICache.
\section{Evaluation}
\begin{figure}[t]
    \centering
    \begin{subfigure}[t]{\columnwidth}
        \includegraphics[width=\textwidth]{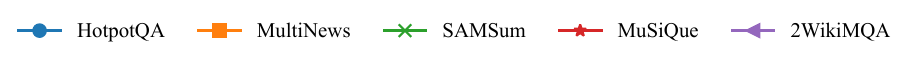}
    \end{subfigure}
    \begin{subfigure}[t]{0.48\columnwidth}
        \includegraphics[width=\textwidth]{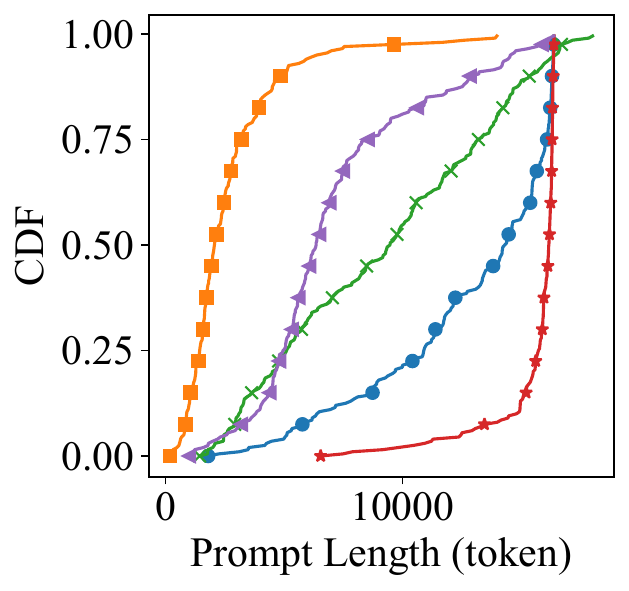}
        \caption{Prompt length distribution}
    \end{subfigure}
    \begin{subfigure}[t]{0.48\columnwidth}
        \includegraphics[width=\textwidth]{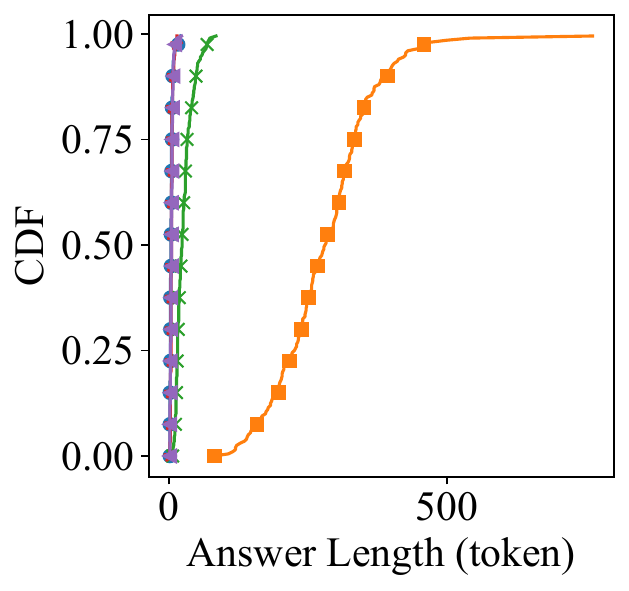}
        \caption{Answer length distribution}
    \end{subfigure}
    \caption{Prefill and decode length distribution.}
    \label{fig-eval-data-dist}
\end{figure}
\label{sec-eval}

We begin by describing the experimental setup, including datasets, models, evaluation metrics, and software/hardware environment. We then present our key evaluation results.

\subsection{Experimental Setup}

\textbf{Dataset.} Following CacheBlend~\cite{yao2025cacheblend} and EPIC~\cite{hu2025epic}, we evaluate on the following LongBench datasets~\cite{bai2024longbench}: \textit{2WikiMQA} \& \textit{HotpotQA} (multi-document question answering), \textit{MuSiQue} (long-document question answering), \textit{SAMSum} (few-shot instruction following), and \textit{MultiNews} (multi-document summarization). 
All datasets contain 200 test cases, with the distribution of prompt (prefill) lengths and answer (decode) lengths shown in Figure~\ref{fig-eval-data-dist}. Static tokens constitute approximately 95\%-99\% of the prompt, while dynamic tokens are fewer than 50.

\textbf{Metrics.} We use the following three metrics to evaluate performance and model accuracy. First, \textit{Time-To-First-Token (TTFT)}~\cite{kwon2023vllm} (lower is better) is used to evaluate all datasets. This metric measures the prefill-stage time: the time from when users send a request to when users receive the first token; this time could be reduced by using context caching. Second, \textit{F1 score}~\cite{bai2024longbench} (higher is better) is used to evaluate \textit{2WikiMQA}, \textit{MuSiQue}, and \textit{HotpotQA}. This metric measures the similarity between LLMs' output and the ground-truth answer based on their common words. Third, \textit{Rough-L score}~\cite{lin2004rouge} (higher is better) is used to evaluate \textit{SAMSum} and \textit{MultiNews}. This metric measures the similarity between LLMs' output and the ground-truth answer by calculating the length of their longest common subsequence.


\begin{figure}
    \centering
    \begin{subfigure}[t]{\columnwidth}
        \includegraphics[width=\textwidth]{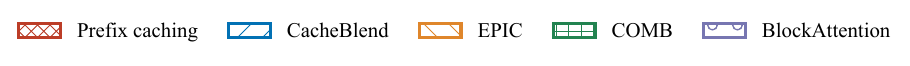}
    \end{subfigure}
    \begin{subfigure}[t]{\columnwidth}
        \includegraphics[width=\textwidth]{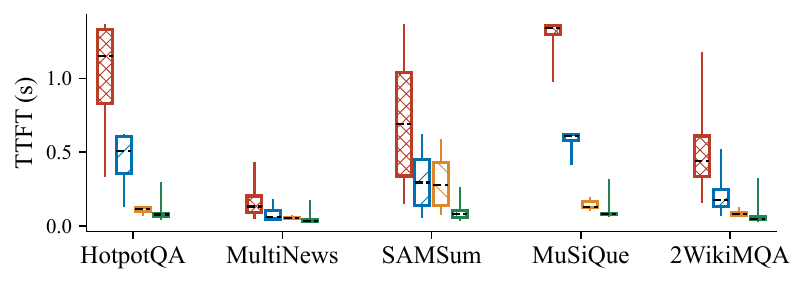}
        \caption{TTFT upon cache hit}
        \label{fig:llama-ttft}
    \end{subfigure}
    \begin{subfigure}[t]{0.55\columnwidth}
        \includegraphics[width=\textwidth]{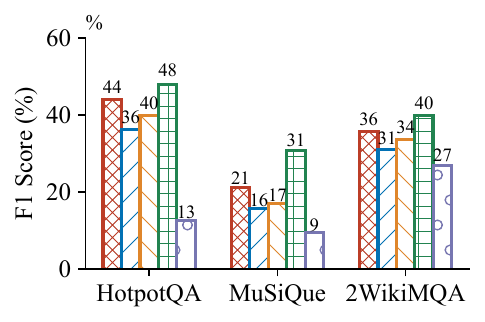}
        \caption{F1 score}
        \label{fig:llama-f1}
    \end{subfigure}
    \begin{subfigure}[t]{0.4\columnwidth}
        \includegraphics[width=\textwidth]{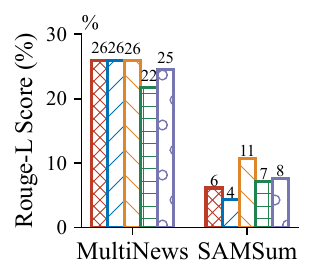}
        \caption{RougeL score}
        \label{fig:llama-RougeL}
    \end{subfigure}
    \caption{Comparison of TTFT ($\downarrow$ Better) and accuracy ($\uparrow$ Better) using  Llama-3.1-8B-Instruct. Note that the TTFT of BlockAttention is not shown because it is not integrated with vLLM currently.}
\end{figure}

\textbf{Baselines.} We compare \sys\ against four representative approaches: Prefix caching, CacheBlend~\cite{yao2025cacheblend}, EPIC~\cite{hu2025epic}, and BlockAttention~\cite{ma2025blockattention}. Prefix caching follows the standard attention mechanism; EPIC and CacheBlend are post-training PIC approaches, while BlockAttention is a training-aware PIC approach. We set the recomputation ratio of CacheBlend to 20\%, and set the number of recomputed tokens per document to 64 for EPIC.

\textbf{Environment.} We run experiments on a single NVIDIA A100 server with four A100-80GB GPU available. The server has 128-core Intel(R) Xeon(R) Platinum 8358P CPU@2.60GHz with 2 hyperthreading and 1TB DRAM. We use Ubuntu 20.04 with Linux kernel 5.16.7 and CUDA 12.6. The Inference engine that we use for evaluations is vLLM 0.12.0.

\subsection{End-to-End Accuracy under PIC}

We first examine whether \sys\ preserves the generation quality of the underlying decoder-only models when PIC is enabled.
Figures~\ref{fig:llama-f1} and~\ref{fig:llama-RougeL} report F1 and Rouge-L scores for Llama-3.1-8B-Instruct, and Figures~\ref{fig:deepseek-f1} and~\ref{fig:deepseek-RougeL} report the same metrics for DeepSeek-V2-Lite-Chat.
Across all datasets and both model families, \sys\ attains high accuracy among the evaluated PIC-based approaches.

We do not treat absolute accuracy as a primary contribution, as the comparison is not fully controlled. In particular, \sys\ is explicitly trained to be PIC-aware, whereas EPIC is training-free, and other baselines fine-tune fewer parameters for fewer steps. Rather, these results substantiate our central claim: in PIC usage scenarios, incorporating a PIC-aware component into the model architecture and training it accordingly enables the model to recover---and in some cases exceed---the accuracy of standard prefix-based attention. The strong and consistent performance of \sys\ across datasets and across two architecturally distinct model families (Llama and DeepSeek with MLA) thus provides empirical evidence for the effectiveness of PIC-aware training.

Two additional observations are worth noting. First, despite being training-free, EPIC achieves unexpectedly strong performance on the \textit{SAMSum} dataset, in some cases surpassing the prefix-caching baseline. This behavior has been analyzed in detail in the EPIC study~\cite{hu2025epic}, and our results are consistent with their findings. Second, \textsc{COMBDeepSeek} attains substantially higher accuracy than its underlying prefix-caching baseline because the base DeepSeek model produces lower-quality outputs; using higher-quality responses generated by Llama as supervision during training (as discussed in Section~\ref{sec-impl}) leads to a significant accuracy improvement.

\subsection{Non-Intrusiveness of \sys}

\begin{figure}
    \centering
    \begin{subfigure}[t]{\columnwidth}
        \includegraphics[width=\textwidth]{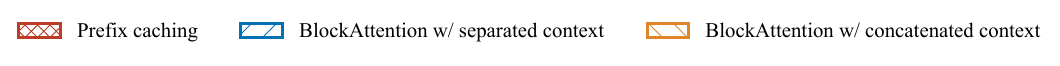}
    \end{subfigure}
    \begin{subfigure}[t]{0.55\columnwidth}
        \includegraphics[width=\textwidth]{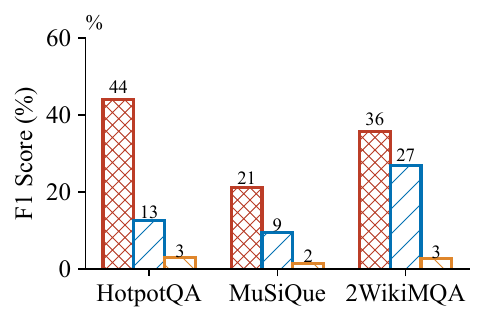}
        \caption{F1 score}
        \label{fig:block-f1}
    \end{subfigure}
    \begin{subfigure}[t]{0.4\columnwidth}
        \includegraphics[width=\textwidth]{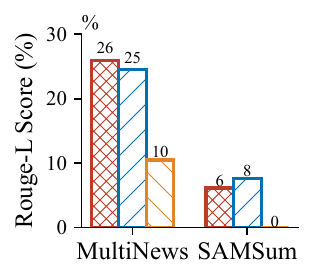}
        \caption{RougeL score}
        \label{fig:block-RougeL}
    \end{subfigure}
    \caption{Comparison of F1 score and RougeL score between using separated context as input to BlockAttention and not using it.}
    \label{fig:block}
\end{figure}

We emphasize that COMB is non-intrusive. In practical or industrial settings, the encoder can be enabled or disabled on demand. When the encoder is disabled, COMB reduces exactly to the underlying decoder-only model and recovers its original prefix-caching behavior and accuracy, without any degradation (just as the prefix caching shown in \figurename~\ref{fig:block}).

In contrast, approaches such as BlockAttention are inherently intrusive. When PIC is not desired, and multiple documents are directly fed into such models, accuracy can collapse. The reason is that BlockAttention is explicitly trained to restrict attention within predefined document blocks. If users do not carefully chunk inputs and apply block attention in its intended manner, the model's assumptions are violated, leading to severe performance degradation (Figure~\ref{fig:block-f1} and Figure~\ref{fig:block-RougeL}, blue bars vs. orange bars).

\subsection{TTFT under Cache Hits and Misses}

\begin{figure}
    \centering
    \begin{subfigure}[t]{\columnwidth}
        \includegraphics[width=\textwidth]{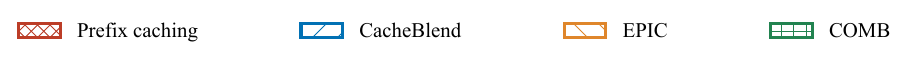}
    \end{subfigure}
    \begin{subfigure}[t]{\columnwidth}
        \includegraphics[width=\textwidth]{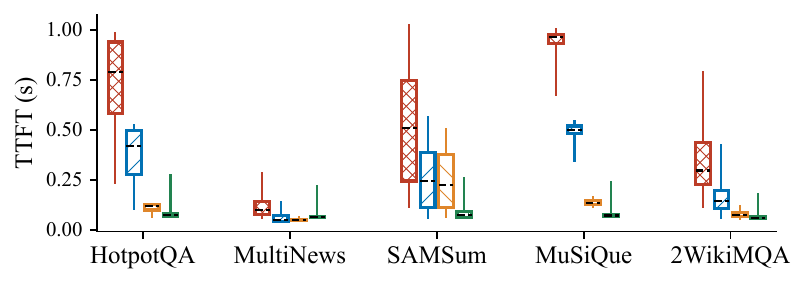}
        \caption{TTFT upon cache hit}
        \label{fig:deepseek-ttft}
    \end{subfigure}
    \begin{subfigure}[t]{0.55\columnwidth}
        \includegraphics[width=\textwidth]{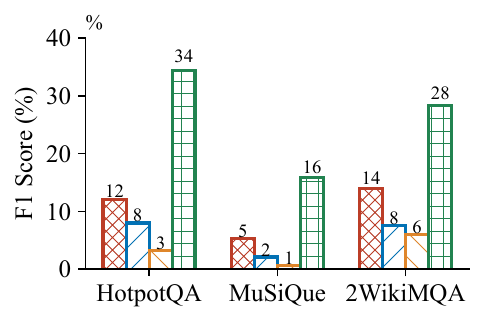}
        \caption{F1 score}
        \label{fig:deepseek-f1}
    \end{subfigure}
    \begin{subfigure}[t]{0.4\columnwidth}
        \includegraphics[width=\textwidth]{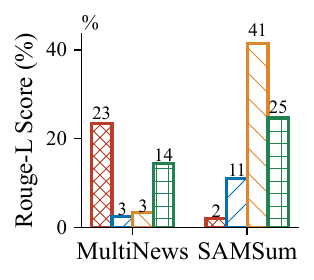}
        \caption{RougeL score}
        \label{fig:deepseek-RougeL}
    \end{subfigure}
    \caption{Comparison of TTFT ($\downarrow$ Better) and accuracy ($\uparrow$ Better) using  DeepSeek-V2-Lite-Chat. Note that BlockAttention is not shown because it was not trained on DeepSeek-V2-Lite-Chat.}
\end{figure}

\begin{figure}
    \centering
    \begin{subfigure}[t]{0.6\columnwidth}
        \includegraphics[width=\textwidth]{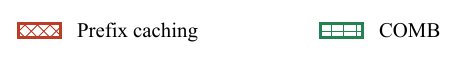}
    \end{subfigure}
    \begin{minipage}{0.45\columnwidth}
        \centering
        \includegraphics[width=\textwidth]{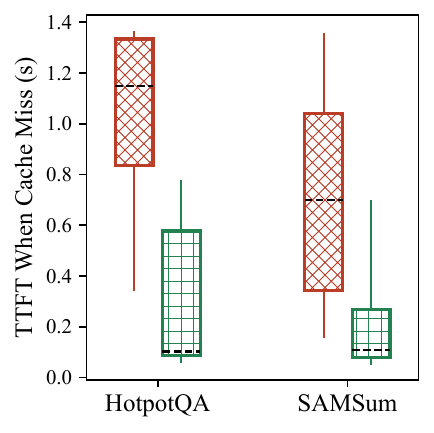}
        \caption{TTFT without KV cache using Llama-3.1-8B.}
        \label{fig:cache-ttft}
    \end{minipage}
    \hfill
    \begin{minipage}{0.5\columnwidth}
        \centering
        \includegraphics[width=\textwidth]{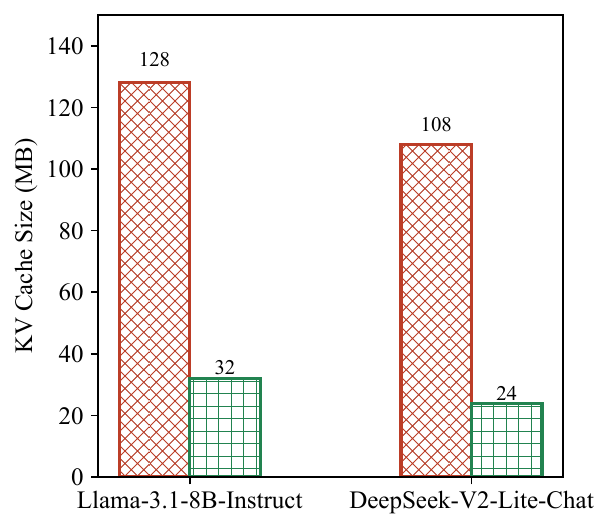}
        \caption{The KV Cache Size of 1,024 tokens.}
        \label{fig:cache-size}
    \end{minipage}
\end{figure}

We next analyze TTFT with and without PICaches. For each sample in the test set, we first feed it into the inference system and record the TTFT for cache misses as shown in \figurename~\ref{fig:cache-ttft}. Then, we modify the sample's instruction to simulate another request with the same context but different prompt, thereby recording the TTFT for cache hits as shown in Figures~\ref{fig:llama-ttft} and~\ref{fig:deepseek-ttft}.
As expected, all PIC approaches substantially reduce TTFT under cache hits, since most document-side computation is amortized across requests.
Among existing PIC approaches, EPIC attains particularly low TTFT because it recomputes only several tokens at each chunk boundary during the link stage, incurring minimal marginal overhead.
Training-aware approaches such as BlockAttention offer algorithmic advantages, but in practice have not been integrated into production-grade inference engines (e.g., vLLM), and thus are typically evaluated only with vanilla HuggingFace runtime. Therefore, the TTFT of BlockAttention is significantly higher than other approaches, and we do not show it in the figures.

\sys\ achieves the lowest TTFT due to the lightweight encoder architecture and the time complexity analyzed in Section~\ref{sec-model}: encoder self-attention is applied only to document tokens, decoder self-attention only to query tokens, and cross-attention couples them in $O(l_{\text{query}} \, l_{\text{doc}})$ time.
PICaches for each document are compiled once and reused across subsequent requests, so when caches hit, the decoder only needs to mix relatively short query sequences with precomputed document PICaches via cross-attention.

When caches miss, \sys\ performs a cold-start compilation to generate PICaches for new documents. Even in this regime, TTFT remains competitive with or better than the decoder-only baseline for two reasons. First, the encoder is substantially shallower than the decoder and is applied only to static document tokens, so the additional prefill cost is modest. Second, production-level inference engines such as vLLM already employ chunked prefill: instead of processing extremely long prompts in a single pass, they split inputs into manageable segments and prefill them incrementally to avoid large intermediate activations and out-of-memory errors.
PIC compilation in \sys\ naturally aligns with this chunked prefill behavior, so much of the cold-start work is overlapped with computation that the baseline system would perform regardless.
As a result, \sys\ provides low TTFT in both cache-hit and cache-miss scenarios while enabling accurate, model-aware PIC.

\subsection{Online Latency and Throughput under Increasing Load}

\begin{figure}
    \centering
    \begin{subfigure}[t]{\columnwidth}
        \includegraphics[width=\textwidth]{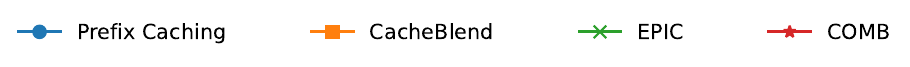}
    \end{subfigure}
    \begin{subfigure}[t]{0.50\columnwidth}
        \includegraphics[width=\textwidth]{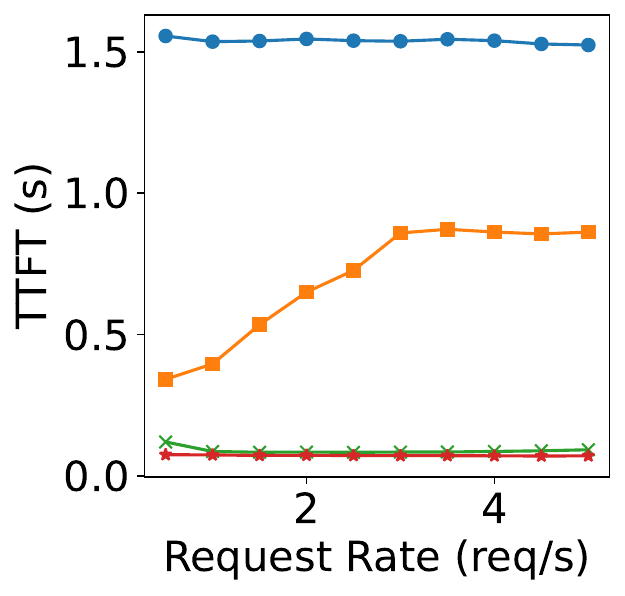}
        \caption{TTFT}
        \label{fig:online-a}
    \end{subfigure}
    \begin{subfigure}[t]{0.47\columnwidth}
        \includegraphics[width=\textwidth]{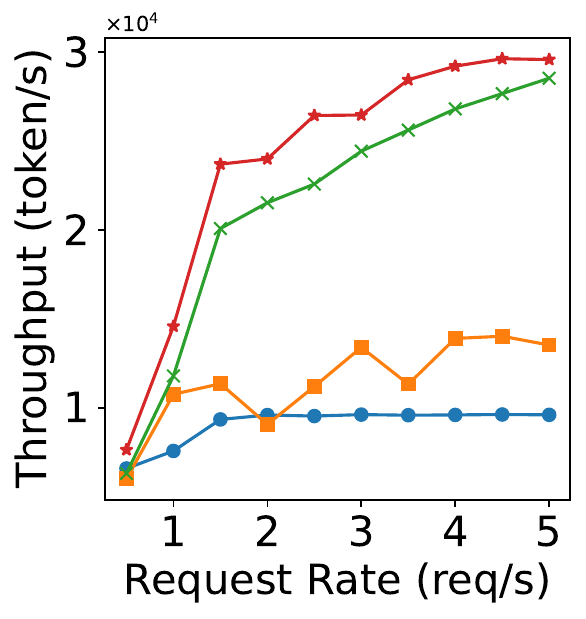}
        \caption{Throughput}
        \label{fig:online-b}
    \end{subfigure}
    \caption{The performance of \sys\ as the request rate increases. }
\end{figure}

Finally, we evaluate \sys\ under increasing request rates in an online serving setting.
Figures~\ref{fig:online-a} and~\ref{fig:online-b} report TTFT and throughput, respectively, as the number of concurrent users and, correspondingly, the amount of GPU memory used by position-independent caches increase.
Across the evaluated load range, \sys\ maintains the lowest TTFT while simultaneously achieving the highest sustained throughput among all approaches.

These gains are closely tied to the memory complexity in Section~\ref{sec-model}.
In a standard decoder-only architecture, KV memory scales with the \emph{entire} prompt length (documents plus query), quickly exhausting GPU HBM and limiting concurrency.
By contrast, \sys\ stores encoder KVs only for static document tokens across a small number of encoder layers, and decoder KVs only for the much shorter query sequence.
This lightweight encoder design reduces per-request KV memory, freeing HBM capacity that can instead be used to batch more requests or admit more users.
As shown in \figurename~\ref{fig:cache-size}, \sys\ saves KV memory by 75\% for Llama-3.1-8B-Instruct, and by 78\% for DeepSeek-V2-Lite.

Taken together with the accuracy results above, these findings show that native, PIC-aware training delivers a favorable accuracy--latency--throughput trade-off in realistic serving scenarios.

\section{Discussion}
\subsection{Why does \sys\ Work?}
Our journey began with optimizing PIC accuracy, driven by the goal of integrating PIC as a native architectural feature rather than an afterthought. This led us to an encoder-decoder design leveraging cross-attention—a conviction we held firmly even before committing substantial resources to large-scale training.

This approach rests on two foundational observations. First, the Transformer architecture was originally conceived as an encoder-decoder framework~\cite{vaswani2017attention}. The encoder's inherent strength in comprehension tasks makes it ideally suited to process arbitrarily ordered contexts, perfectly complementing the decoder's generation capabilities. Second, this paradigm has already been validated in multimodal LLMs such as Whisper~\cite{radford2023whisper} and Mllama~\cite{meta2024llama32}. Multimodal tasks are inherently more challenging than pure language tasks—requiring the interpretation of text potentially embedded within images and complex visual reasoning. If this architecture proves effective for such demanding multimodal scenarios, its applicability to pure language modeling follows naturally.

\subsection{The Potential of PIC}
We have now entered the era of AI Agents. What is an Agent, and how does it differ from an LLM? A widely accepted view holds that Agents can formulate short-term actions based on long-term goals and are capable of using tools. Consequently, \emph{retrieval} emerges as a crucial capability for Agents. Beyond consulting references to gather information, Agents must also retrieve which tools are best suited for specific tasks.

Whenever retrieval is involved, PIC becomes essential, because retrieved items can arrive in \emph{arbitrary order} under various permutations and combinations. Using prefix caching in this scenario proves highly inefficient: essentially only the first item can reuse the KV cache, while all subsequent items fail to do so.

We can address this by equipping the Agent with an \emph{Encoder}: all retrieved content—arriving in arbitrary order—is fed into the Encoder, while the question and the model's reasoning remain in the Decoder. When the model needs to retrieve new information, previously retrieved content is discarded and replaced with new items in the Encoder.

Imagine how powerful such an Agent would be if the Decoder's 128K context window contained only the question and the model's own chain of thought, free from noisy reference materials.

\section{Related Work}
\label{sec-related}

Our work is unique in proposing native position-independent caching. In this section, we briefly navigate the whole design space.

\paragraph{LLM Serving Optimizations.} Several serving systems have emerged in recent years. vLLM~\cite{kwon2023vllm} is a pioneering work in this space featuring PagedAttention for higher throughput. SGLang~\cite{zheng2024sglang} is another serving system featuring a novel frontend language and a backend runtime. Aside from full systems, there are also many scheduling optimizations, such as disaggregated prefill and decode~\cite{zhong2024distserve, qin2024mooncake, patel2024splitwise}, continuous batching~\cite{yu2022orca}, speculative decoding~\cite{miao2024specinfer}, etc.

\paragraph{Context Caching (CC).} CC has two main categories. The first is position-dependent caching, which can be further divided into prefix-based caching~\cite{yu2023stateful-pensieve, liu2023cachegen, zheng2024sglang} and relative position-dependent caching, such as PromptCache~\cite{gim2024prompt}. By mid-2024, vendors such as Gemini~\cite{gemini-context-caching} and DeepSeek~\cite{deepseek_api} began incorporating explicit prefix-based CC features into their systems. The second category is Position-Independent Caching (PIC)~\cite{hu2025epic, yao2025cacheblend, lu2025turborag, ma2025blockattention}. Although EPIC~\cite{hu2025epic} and CacheBlend~\cite{yao2025cacheblend} address initial aspects of PIC, \sys\ advances the paradigm by rendering PIC intrinsic to the model architecture.

\paragraph{Post-Training vs. Training-Aware Approaches.} A recurring trajectory in LLM research involves initially validating hypotheses via post-training adjustments before integrating them into the model training process. For instance, in the realm of attention sparsity, researchers first explored inference-time heuristics such as H2O~\cite{zhang2023h2o}, Quest~\cite{tang2024quest}, and RaaS~\cite{hu2025raas}. Following the validation of these sparsity patterns~\cite{jiang2024minference}, DeepSeek introduced Native Sparse Attention (NSA)~\cite{nsa2025yuan} to incorporate sparse structures directly into the model. This progression from post-training experimentation to training-aware design is also evident in quantization~\cite{xiao2023smoothquant, lin2024awq}, pruning~\cite{wang2023ntksap, zhang2021lottery}, speculative decoding~\cite{deepseekai2025deepseekv3}, and attention sink~\cite{qiu2025gated}. The underlying principle is that once an inductive bias---whether sparsity or position-independent caching---proves beneficial post-training, it merits integration into the model's native architecture. While EPIC~\cite{hu2025epic} and CacheBlend~\cite{yao2025cacheblend} represent post-training approaches, BlockAttention~\cite{ma2025blockattention}, KVLink~\cite{yang2025kvlink}, TurboRAG~\cite{lu2025turborag}, and \sys\ fall under the training-aware category. Notably, \sys\ combines the best of both worlds, achieving the accuracy benefits of training-aware approaches while retaining the non-intrusive, plug-in flexibility of post-training approaches.
\section{Conclusion}\label{sec-conclusion}

In this paper, we have proposed \textbf{native PIC} by reintroducing an encoder into decoder-only LLMs and explicitly training it for PIC-compatible generation. We have also presented \sys, a PIC-aware caching system integrated with existing inference frameworks. Extensive experiments demonstrate that \sys\ achieves the highest accuracy under PIC while substantially improving inference efficiency, reducing TTFT by up to 94\% and increasing throughput by up to 3$\times$. In addition, \sys\ remains a plug-in mechanism that can be enabled or disabled on demand, reverting to standard prefix caching when PIC is not used.
\bibliographystyle{plain}
\bibliography{references}

\begin{thebibliography}{10}

\bibitem{agarwal2025cachecraft}
Shubham Agarwal, Sai Sundaresan, Subrata Mitra, Debabrata Mahapatra, Archit Gupta, Rounak Sharma, Nirmal~Joshua Kapu, Tong Yu, and Shiv Saini.
\newblock Cache-craft: Managing chunk-caches for efficient retrieval-augmented generation.
\newblock {\em Proc. ACM Manag. Data}, 3(3), June 2025.

\bibitem{bai2024longbench}
Yushi Bai, Xin Lv, Jiajie Zhang, Hongchang Lyu, Jiankai Tang, Zhidian Huang, Zhengxiao Du, Xiao Liu, Aohan Zeng, Lei Hou, Yuxiao Dong, Jie Tang, and Juanzi Li.
\newblock {LongBench}: {A} bilingual, multitask benchmark for long context understanding.
\newblock In {\em Proceedings of the Sixty-Second Annual Meeting of the Association for Computational Linguistics}, pages 3119--3137, 2024.

\bibitem{deepseek_api}
Deepseek.
\newblock Deepseek api introduces context caching on disk, cutting prices by an order of magnitude.
\newblock \url{https://api-docs.deepseek.com/news/news0802}, 2024.

\bibitem{deepseekai2025deepseekv3}
DeepSeek-AI, Aixin Liu, Bei Feng, Bing Xue, Bingxuan Wang, Bochao Wu, Chengda Lu, Chenggang Zhao, Chengqi Deng, Chenyu Zhang, Chong Ruan, Damai Dai, Daya Guo, Dejian Yang, Deli Chen, Dongjie Ji, Erhang Li, Fangyun Lin, Fucong Dai, Fuli Luo, Guangbo Hao, Guanting Chen, Guowei Li, H.~Zhang, Han Bao, Hanwei Xu, Haocheng Wang, Haowei Zhang, Honghui Ding, Huajian Xin, Huazuo Gao, Hui Li, Hui Qu, J.~L. Cai, Jian Liang, Jianzhong Guo, Jiaqi Ni, Jiashi Li, Jiawei Wang, Jin Chen, Jingchang Chen, Jingyang Yuan, Junjie Qiu, Junlong Li, Junxiao Song, Kai Dong, Kai Hu, Kaige Gao, Kang Guan, Kexin Huang, Kuai Yu, Lean Wang, Lecong Zhang, Lei Xu, Leyi Xia, Liang Zhao, Litong Wang, Liyue Zhang, Meng Li, Miaojun Wang, Mingchuan Zhang, Minghua Zhang, Minghui Tang, Mingming Li, Ning Tian, Panpan Huang, Peiyi Wang, Peng Zhang, Qiancheng Wang, Qihao Zhu, Qinyu Chen, Qiushi Du, R.~J. Chen, R.~L. Jin, Ruiqi Ge, Ruisong Zhang, Ruizhe Pan, Runji Wang, Runxin Xu, Ruoyu Zhang, Ruyi Chen, S.~S. Li, Shanghao Lu, Shangyan Zhou, Shanhuang
  Chen, Shaoqing Wu, Shengfeng Ye, Shengfeng Ye, Shirong Ma, Shiyu Wang, Shuang Zhou, Shuiping Yu, Shunfeng Zhou, Shuting Pan, T.~Wang, Tao Yun, Tian Pei, Tianyu Sun, W.~L. Xiao, Wangding Zeng, Wanjia Zhao, Wei An, Wen Liu, Wenfeng Liang, Wenjun Gao, Wenqin Yu, Wentao Zhang, X.~Q. Li, Xiangyue Jin, Xianzu Wang, Xiao Bi, Xiaodong Liu, Xiaohan Wang, Xiaojin Shen, Xiaokang Chen, Xiaokang Zhang, Xiaosha Chen, Xiaotao Nie, Xiaowen Sun, Xiaoxiang Wang, Xin Cheng, Xin Liu, Xin Xie, Xingchao Liu, Xingkai Yu, Xinnan Song, Xinxia Shan, Xinyi Zhou, Xinyu Yang, Xinyuan Li, Xuecheng Su, Xuheng Lin, Y.~K. Li, Y.~Q. Wang, Y.~X. Wei, Y.~X. Zhu, Yang Zhang, Yanhong Xu, Yanhong Xu, Yanping Huang, Yao Li, Yao Zhao, Yaofeng Sun, Yaohui Li, Yaohui Wang, Yi~Yu, Yi~Zheng, Yichao Zhang, Yifan Shi, Yiliang Xiong, Ying He, Ying Tang, Yishi Piao, Yisong Wang, Yixuan Tan, Yiyang Ma, Yiyuan Liu, Yongqiang Guo, Yu~Wu, Yuan Ou, Yuchen Zhu, Yuduan Wang, Yue Gong, Yuheng Zou, Yujia He, Yukun Zha, Yunfan Xiong, Yunxian Ma, Yuting Yan, Yuxiang
  Luo, Yuxiang You, Yuxuan Liu, Yuyang Zhou, Z.~F. Wu, Z.~Z. Ren, Zehui Ren, Zhangli Sha, Zhe Fu, Zhean Xu, Zhen Huang, Zhen Zhang, Zhenda Xie, Zhengyan Zhang, Zhewen Hao, Zhibin Gou, Zhicheng Ma, Zhigang Yan, Zhihong Shao, Zhipeng Xu, Zhiyu Wu, Zhongyu Zhang, Zhuoshu Li, Zihui Gu, Zijia Zhu, Zijun Liu, Zilin Li, Ziwei Xie, Ziyang Song, Ziyi Gao, and Zizheng Pan.
\newblock Deepseek-v3 technical report, 2025.

\bibitem{gim2024prompt}
In~Gim, Guojun Chen, Seung-seob Lee, Nikhil Sarda, Anurag Khandelwal, and Lin Zhong.
\newblock Prompt cache: Modular attention reuse for low-latency inference.
\newblock {\em Proceedings of Machine Learning and Systems}, 6:325--338, 2024.

\bibitem{gemini-context-caching}
Google.
\newblock Gemini context caching.
\newblock \url{https://ai.google.dev/gemini-api/docs/caching}, 2026.

\bibitem{hu2025epic}
Junhao Hu, Wenrui Huang, Haoyi Wang, Weidong Wang, Tiancheng Hu, Qin Zhang, Hao Feng, Xusheng Chen, Yizhou Shan, and Tao Xie.
\newblock {EPIC:} efficient position-independent caching for serving large language models.
\newblock In {\em Proceedings of the 42nd International Conference on Machine Learning}, pages 24391--24402, 2025.

\bibitem{hu2025raas}
Junhao Hu, Wenrui Huang, Weidong Wang, Zhenwen Li, Tiancheng Hu, Zhixia Liu, Xusheng Chen, Tao Xie, and Yizhou Shan.
\newblock {RaaS}: Reasoning-aware attention sparsity for efficient llm reasoning.
\newblock In {\em Proceedings of the 63rd Annual Meeting of the Association for Computational Linguistics}, pages 2577--2590, 2024.

\bibitem{hu2023pcrml}
Junhao Hu, Chaozheng Wang, Hailiang Huang, Huang Luo, Yu~Jin, Yuetang Deng, and Tao Xie.
\newblock Predicting compilation resources for adaptive build in an industrial setting.
\newblock In {\em Proceedings of the 38th {IEEE/ACM} International Conference on Automated Software Engineering}, pages 1808--1813, 2023.

\bibitem{jiang2024minference}
Huiqiang Jiang, Yucheng Li, Chengruidong Zhang, Qianhui Wu, Xufang Luo, Surin Ahn, Zhenhua Han, Amir~H Abdi, Dongsheng Li, Chin-Yew Lin, et~al.
\newblock Minference 1.0: Accelerating pre-filling for long-context llms via dynamic sparse attention.
\newblock {\em arXiv preprint arXiv:2407.02490}, 2024.

\bibitem{kwon2023vllm}
Woosuk Kwon, Zhuohan Li, Siyuan Zhuang, Ying Sheng, Lianmin Zheng, Cody~Hao Yu, Joseph~E. Gonzalez, Hao Zhang, and Ion Stoica.
\newblock Efficient memory management for large language model serving with pagedattention.
\newblock In {\em Proceedings of the ACM SIGOPS 29th Symposium on Operating Systems Principles}, 2023.

\bibitem{liang2025diff}
Qingyuan Liang, Zeyu Sun, Qihao Zhu, Junhao Hu, Yifan Zhao, Yizhou Chen, Mingxuan Zhu, Guoqing Wang, and Lu~Zhang.
\newblock Directional diffusion-style code editing pre-training.
\newblock {\em {IEEE} Trans. Software Eng.}, 51(9):2583--2600, 2025.

\bibitem{lin2004rouge}
Chin-Yew Lin.
\newblock Rouge: A package for automatic evaluation of summaries.
\newblock In {\em Text summarization branches out}, pages 74--81, 2004.

\bibitem{lin2024awq}
Ji~Lin, Jiaming Tang, Haotian Tang, Shang Yang, Wei{-}Ming Chen, Wei{-}Chen Wang, Guangxuan Xiao, Xingyu Dang, Chuang Gan, and Song Han.
\newblock {AWQ:} activation-aware weight quantization for on-device {LLM} compression and acceleration.
\newblock In {\em Proceedings of the 7th Annual Conference on Machine Learning and Systems}, 2024.

\bibitem{liu2023cachegen}
Yuhan Liu, Hanchen Li, Kuntai Du, Jiayi Yao, Yihua Cheng, Yuyang Huang, Shan Lu, Michael Maire, Henry Hoffmann, Ari Holtzman, et~al.
\newblock Cachegen: Fast context loading for language model applications.
\newblock {\em arXiv preprint arXiv:2310.07240}, 2023.

\bibitem{lu2025turborag}
Songshuo Lu, Hua Wang, Yutian Rong, Zhi Chen, and Yaohua Tang.
\newblock {T}urbo{RAG}: Accelerating retrieval-augmented generation with precomputed {KV} caches for chunked text.
\newblock In Christos Christodoulopoulos, Tanmoy Chakraborty, Carolyn Rose, and Violet Peng, editors, {\em Proceedings of the 2025 Conference on Empirical Methods in Natural Language Processing}, pages 6588--6601, Suzhou, China, November 2025. Association for Computational Linguistics.

\bibitem{ma2025blockattention}
Dongyang Ma, Yan Wang, and Tian Lan.
\newblock Block-attention for efficient prefilling.
\newblock In {\em The Thirteenth International Conference on Learning Representations}, 2025.

\bibitem{llama3.1}
{Meta}.
\newblock {Introducing Llama 3.1: Our most capable models to date}.
\newblock \url{https://ai.meta.com/blog/meta-llama-3-1/}.

\bibitem{meta2024llama32}
{Meta AI}.
\newblock Llama 3.2: Revolutionizing edge {AI} and vision with open, customizable models, sep 2024.

\bibitem{miao2024specinfer}
Xupeng Miao, Gabriele Oliaro, Zhihao Zhang, Xinhao Cheng, Zeyu Wang, Zhengxin Zhang, Rae Ying~Yee Wong, Alan Zhu, Lijie Yang, Xiaoxiang Shi, Chunan Shi, Zhuoming Chen, Daiyaan Arfeen, Reyna Abhyankar, and Zhihao Jia.
\newblock Specinfer: Accelerating large language model serving with tree-based speculative inference and verification.
\newblock In {\em Proceedings of the 29th ACM International Conference on Architectural Support for Programming Languages and Operating Systems, Volume 3}, ASPLOS '24, page 932–949, New York, NY, USA, 2024. Association for Computing Machinery.

\bibitem{mishra2022naturalinstructions}
Swaroop Mishra, Daniel Khashabi, Chitta Baral, and Hannaneh Hajishirzi.
\newblock Cross-task generalization via natural language crowdsourcing instructions.
\newblock In {\em ACL}, 2022.

\bibitem{narayan2018dont}
Shashi Narayan, Shay~B. Cohen, and Mirella Lapata.
\newblock Don{'}t give me the details, just the summary! topic-aware convolutional neural networks for extreme summarization.
\newblock In Ellen Riloff, David Chiang, Julia Hockenmaier, and Jun{'}ichi Tsujii, editors, {\em Proceedings of the 2018 Conference on Empirical Methods in Natural Language Processing}, pages 1797--1807, Brussels, Belgium, October-November 2018. Association for Computational Linguistics.

\bibitem{patel2024splitwise}
Pratyush Patel, Esha Choukse, Chaojie Zhang, Aashaka Shah, \'{I}ñigo Goiri, Saeed Maleki, and Ricardo Bianchini.
\newblock Splitwise: Efficient generative llm inference using phase splitting.
\newblock In {\em 2024 ACM/IEEE 51st Annual International Symposium on Computer Architecture (ISCA)}, pages 118--132, 2024.

\bibitem{qin2024mooncake}
Ruoyu Qin, Zheming Li, Weiran He, Jialei Cui, Feng Ren, Mingxing Zhang, Yongwei Wu, Weimin Zheng, and Xinran Xu.
\newblock Mooncake: Trading more storage for less computation - {A} {KVCache}-centric architecture for serving {LLM} chatbot.
\newblock In {\em Proceedings of the 23rd {USENIX} Conference on File and Storage Technologies}, pages 155--170, 2025.

\bibitem{qiu2025gated}
Zihan Qiu, Zekun Wang, Bo~Zheng, Zeyu Huang, Kaiyue Wen, Songlin Yang, Rui Men, Le~Yu, Fei Huang, Suozhi Huang, Dayiheng Liu, Jingren Zhou, and Junyang Lin.
\newblock Gated attention for large language models: Non-linearity, sparsity, and attention-sink-free.
\newblock In {\em The Thirty-ninth Annual Conference on Neural Information Processing Systems}, 2025.

\bibitem{radford2023whisper}
Alec Radford, Jong~Wook Kim, Tao Xu, Greg Brockman, Christine McLeavey, and Ilya Sutskever.
\newblock Robust speech recognition via large-scale weak supervision.
\newblock In {\em Proceedings of the 40th International Conference on Machine Learning}, ICML'23. JMLR.org, 2023.

\bibitem{rajpurkar2018know}
Pranav Rajpurkar, Robin Jia, and Percy Liang.
\newblock Know what you don{'}t know: Unanswerable questions for {SQ}u{AD}.
\newblock In Iryna Gurevych and Yusuke Miyao, editors, {\em Proceedings of the 56th Annual Meeting of the Association for Computational Linguistics (Volume 2: Short Papers)}, pages 784--789, Melbourne, Australia, July 2018. Association for Computational Linguistics.

\bibitem{tang2024quest}
Jiaming Tang, Yilong Zhao, Kan Zhu, Guangxuan Xiao, Baris Kasikci, and Song Han.
\newblock {QUEST:} query-aware sparsity for efficient long-context {LLM} inference.
\newblock In {\em Proceedings of the 41st International Conference on Machine Learning}, pages 47901--47911, 2024.

\bibitem{vaswani2017attention}
Ashish Vaswani, Noam Shazeer, Niki Parmar, Jakob Uszkoreit, Llion Jones, Aidan~N. Gomez, \L{}ukasz Kaiser, and Illia Polosukhin.
\newblock Attention is all you need.
\newblock In {\em Proceedings of the 31st International Conference on Neural Information Processing Systems}, NIPS'17, page 6000–6010, Red Hook, NY, USA, 2017. Curran Associates Inc.

\bibitem{wang2023how}
Chaozheng Wang, Junhao Hu, Cuiyun Gao, Yu~Jin, Tao Xie, Hailiang Huang, Zhenyu Lei, and Yuetang Deng.
\newblock How practitioners expect code completion?
\newblock In {\em Proceedings of the 31st {ACM} Joint European Software Engineering Conference and Symposium on the Foundations of Software Engineering}, pages 1294--1306, 2023.

\bibitem{wang2023ntksap}
Yite Wang, Dawei Li, and Ruoyu Sun.
\newblock {NTK-SAP:} improving neural network pruning by aligning training dynamics.
\newblock In {\em Proceedings of the 11th International Conference on Learning Representations}, 2023.

\bibitem{Wang2022supernaturalinstructions}
Yizhong Wang, Swaroop Mishra, Pegah Alipoormolabashi, Yeganeh Kordi, Amirreza Mirzaei, Anjana Arunkumar, Arjun Ashok, Arut~Selvan Dhanasekaran, Atharva Naik, David Stap, et~al.
\newblock Super-naturalinstructions:generalization via declarative instructions on 1600+ tasks.
\newblock In {\em EMNLP}, 2022.

\bibitem{xiao2023smoothquant}
Guangxuan Xiao, Ji~Lin, Micka{\"{e}}l Seznec, Hao Wu, Julien Demouth, and Song Han.
\newblock Smoothquant: Accurate and efficient post-training quantization for large language models.
\newblock In {\em Proceedings of the 40th International Conference on Machine Learning}, volume 202, pages 38087--38099, 2023.

\bibitem{yang2025kvlink}
Jingbo Yang, Bairu Hou, Wei Wei, Yujia Bao, and Shiyu Chang.
\newblock {KVL}ink: Accelerating large language models via efficient {KV} cache reuse.
\newblock In {\em The Thirty-ninth Annual Conference on Neural Information Processing Systems}, 2025.

\bibitem{yao2025cacheblend}
Jiayi Yao, Hanchen Li, Yuhan Liu, Siddhant Ray, Yihua Cheng, Qizheng Zhang, Kuntai Du, Shan Lu, and Junchen Jiang.
\newblock Cacheblend: Fast large language model serving for rag with cached knowledge fusion.
\newblock In {\em Proceedings of the Twentieth European Conference on Computer Systems}, EuroSys '25, page 94–109, New York, NY, USA, 2025. Association for Computing Machinery.

\bibitem{yao2023react}
Shunyu Yao, Jeffrey Zhao, Dian Yu, Nan Du, Izhak Shafran, Karthik~R. Narasimhan, and Yuan Cao.
\newblock {ReAct}: Synergizing reasoning and acting in language models.
\newblock In {\em The 11th International Conference on Learning Representations}, 2023.

\bibitem{yu2022orca}
Gyeong-In Yu, Joo~Seong Jeong, Geon-Woo Kim, Soojeong Kim, and Byung-Gon Chun.
\newblock Orca: A distributed serving system for {Transformer-Based} generative models.
\newblock In {\em 16th USENIX Symposium on Operating Systems Design and Implementation (OSDI 22)}, pages 521--538, Carlsbad, CA, July 2022. USENIX Association.

\bibitem{yu2023stateful-pensieve}
Lingfan Yu and Jinyang Li.
\newblock Stateful large language model serving with pensieve.
\newblock {\em arXiv preprint arXiv:2312.05516}, 2023.

\bibitem{nsa2025yuan}
Jingyang Yuan, Huazuo Gao, Damai Dai, Junyu Luo, Liang Zhao, Zhengyan Zhang, Zhenda Xie, Yuxing Wei, Lean Wang, Zhiping Xiao, Yuqing Wang, Chong Ruan, Ming Zhang, Wenfeng Liang, and Wangding Zeng.
\newblock Native sparse attention: Hardware-aligned and natively trainable sparse attention.
\newblock In {\em Proceedings of the 63rd Annual Meeting of the Association for Computational Linguistics}, pages 23078--23097, 2025.

\bibitem{zhang2021lottery}
Zeru Zhang, Jiayin Jin, Zijie Zhang, Yang Zhou, Xin Zhao, Jiaxiang Ren, Ji~Liu, Lingfei Wu, Ruoming Jin, and Dejing Dou.
\newblock Validating the lottery ticket hypothesis with inertial manifold theory.
\newblock In {\em Proceedings of the 35th Annual Conference on Neural Information Processing Systems}, pages 30196--30210, 2021.

\bibitem{zhang2023h2o}
Zhenyu Zhang, Ying Sheng, Tianyi Zhou, Tianlong Chen, Lianmin Zheng, Ruisi Cai, Zhao Song, Yuandong Tian, Christopher R{\'{e}}, Clark~W. Barrett, Zhangyang Wang, and Beidi Chen.
\newblock {H2O:} heavy-hitter oracle for efficient generative inference of large language models.
\newblock In {\em Proceedings of the 37th Annual Conference on Neural Information Processing Systems}, pages 34661--34710, 2023.

\bibitem{zhang2025attention}
Zhisong Zhang, Yan Wang, Xinting Huang, Tianqing Fang, Hongming Zhang, Chenlong Deng, Shuaiyi Li, and Dong Yu.
\newblock Attention entropy is a key factor: An analysis of parallel context encoding with full-attention-based pre-trained language models.
\newblock In Wanxiang Che, Joyce Nabende, Ekaterina Shutova, and Mohammad~Taher Pilehvar, editors, {\em Proceedings of the 63rd Annual Meeting of the Association for Computational Linguistics (Volume 1: Long Papers)}, pages 9840--9855, Vienna, Austria, July 2025. Association for Computational Linguistics.

\bibitem{zhao2025mpic}
Shiju Zhao, Junhao Hu, Rongxiao Huang, Jiaqi Zheng, and Guihai Chen.
\newblock Mpic: Position-independent multimodal context caching system for efficient mllm serving, 2025.

\bibitem{zheng2024sglang}
Lianmin Zheng, Liangsheng Yin, Zhiqiang Xie, Chuyue Sun, Jeff Huang, Cody~Hao Yu, Shiyi Cao, Christos Kozyrakis, Ion Stoica, Joseph~E. Gonzalez, Clark Barrett, and Ying Sheng.
\newblock {SGL}ang: Efficient execution of structured language model programs.
\newblock In {\em The Thirty-eighth Annual Conference on Neural Information Processing Systems}, 2024.

\bibitem{zhong2024distserve}
Yinmin Zhong, Shengyu Liu, Junda Chen, Jianbo Hu, Yibo Zhu, Xuanzhe Liu, Xin Jin, and Hao Zhang.
\newblock {DistServe}: Disaggregating prefill and decoding for goodput-optimized large language model serving.
\newblock In {\em 18th USENIX Symposium on Operating Systems Design and Implementation (OSDI 24)}, pages 193--210, Santa Clara, CA, July 2024. USENIX Association.

\end{thebibliography}
\newpage
\appendix
\end{document}